\documentclass[letterpaper, 10 pt, conference]{ieeeconf}                                             

\usepackage[noadjust,compress,sort,nospace,move]{cite}
\usepackage{hyperref}
\usepackage[export]{adjustbox}
\hypersetup{
	pdftitle={Coupled Particle Filters for Robust Affordance Estimation},
	pdfsubject={Coupled Particle Filters for Robust Affordance Estimation},
	pdfauthor={Patrick Lowin},
	pdfkeywords={}
}
\usepackage{algorithm}
\usepackage{amsmath,amssymb,amsfonts}
\usepackage{algorithmic}
\usepackage{graphicx}
\usepackage{textcomp}
\usepackage{xcolor}
\usepackage{balance}
\def\BibTeX{{\rm B\kern-.05em{\sc i\kern-.025em b}\kern-.08em
    T\kern-.1667em\lower.7ex\hbox{E}\kern-.125emX}}
\usepackage{cleveref}
\crefname{algorithm}{Alg.}{Algs.}
\Crefname{algorithm}{Alg.}{Algs.}
\usepackage{siunitx}

\Crefname{figure}{Fig.}{Figs.}
\Crefname{section}{Sec.}{Secs.}
\Crefname{equation}{Eq.}{Eqs.}

\IEEEoverridecommandlockouts                                                          
\title{\LARGE\bf Coupled Particle Filters for Robust Affordance Estimation}

\author{Patrick Lowin$^{1,2}$ \qquad\ \qquad\ Vito Mengers$^{1,2,3}$ \qquad\ \qquad\ Oliver Brock$^{1,2,3}$
	\thanks{$^1$ Robotics and Biology Laboratory, Technische Universit\"at Berlin}
	\thanks{$^2$ Robotics Institute Germany}
	\thanks{$^3$ Science of Intelligence (SCIoI), Cluster of Excellence, Berlin, Germany}
	\thanks{This work has been partially funded by the German Federal Ministry of Research, Technology and Space (BMFTR) under the Robotics Institute Germany (RIG) and by the Deutsche Forschungsgemeinschaft (DFG, German Research Foundation) under Germany’s Excellence Strategy – EXC 2002 – 390523135.
    }}%
\begin{document}
\maketitle

\begin{abstract}
Robotic affordance estimation is challenging due to visual, geometric, and semantic ambiguities in sensory input. 
We propose a method that disambiguates these signals using two coupled recursive estimators for sub-aspects of affordances: graspable and movable regions. 
Each estimator encodes property-specific regularities to reduce uncertainty, while their coupling enables bidirectional information exchange that focuses attention on regions where both agree, i.e., affordances. 
Evaluated on a real-world dataset, our method outperforms three recent affordance estimators (Where2Act, Hands-as-Probes, and HRP) by 308\%, 245\%, and 257\% in precision, and remains robust under challenging conditions such as low light or cluttered environments. Furthermore, our method achieves a 70\% success rate in our real-world evaluation.
These results demonstrate that coupling complementary estimators yields precise, robust, and embodiment-appropriate affordance predictions.
\end{abstract}

\section{Introduction}
Affordances are the possible actions an agent can perform in its environment~\cite{gibson2014ecological}. For robots that want to interact with the world, detecting affordances is essential yet challenging, as affordance cues are often ambiguous in the sensory input. Since robots can interact by grasping and moving objects, affordances arise in regions where these properties co-occur. In this paper, we present a robust method for detecting affordances by coupling estimators of graspable and movable regions and using their relationship to resolve ambiguities in the sensor~data.

Coupling the estimation processes for graspable and movable regions improves robustness in two ways. First, while we lack explicit affordance priors, we exploit available priors for graspability and movability to reduce uncertainty. For instance, predicted grasps in physically infeasible locations, such as edges caused by depth discontinuities, are rejected. Second, the relationship between the two properties provides an additional prior, allowing the estimators to focus on regions where both agree. This resolves ambiguities that a single estimator cannot.

\begin{figure}[t]
   \centering
    \includegraphics[width=\linewidth]{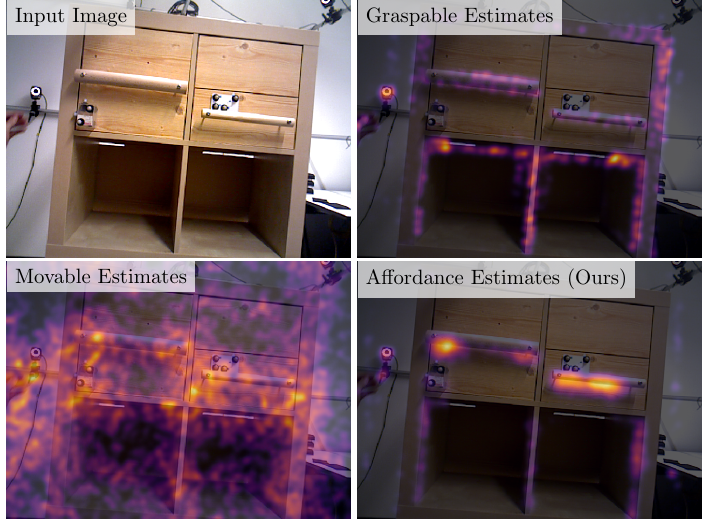}
    \caption{Robots can interact by grasping an object and then moving it. Consequently, affordances emerge in regions where these properties co-occur. By estimating these properties separately and leveraging their relationship, we can resolve ambiguities in both modalities and estimate robust affordances.}
    \label{fig:title}
\end{figure}

Initially, each estimator focuses on the regions most likely for its property. Through information sharing, they gradually align attention on regions important to both, i.e., the true affordances (\Cref{fig:title}). The strength of this coupling can be adjusted to control the interaction between the estimators. By coordinating in this way, they share complementary information, reduce uncertainty, and produce more robust affordance~estimates.

We evaluate our method on the RBO dataset~\cite{rbo-dataset} of articulated objects and interactions, which we extend with manipulable-region annotations. We compare against recent learning-based approaches, outperforming a geometry-based estimator trained in simulation by 308\% and two appearance-based methods trained on human interactions by 245\% and 257\%. At the same time, our approach demonstrates improved robustness under challenging conditions such as low light or clutter. We further evaluate this robustness in real-world manipulation trials, achieving a 70\% success rate. These results highlight that leveraging shared structure, complementary information, and coordinated attention through coupled estimation processes enables robust affordance estimation.

Dataset annotations and code are available at \url{https://github.com/tu-rbo/CoupledParticleFilters}.

\section{Related Work}
Affordance estimation is typically approached by learning affordances in simulation or from human interaction data within a single, monolithic model. We review these approaches in~\Cref{sec:learning_affordances}, highlighting their struggles with out-of-distribution data and perceptual ambiguities. To address these challenges, we instead estimate affordances with two coupled recursive estimators for related properties, an approach that has already proven effective for other complex perceptual problems, as we review in~\Cref{sec:aicon}.

\subsection{Affordance Estimation}\label{sec:learning_affordances}
Affordances are characterized as an agent's possible actions in an environment~\cite{gibson2014ecological}. Their perception thus depends on the agent's capabilities and tasks. Early works used hand-crafted,
often geometry-based rules to predict object affordances~\cite{affordance-tool-detection, visual-affordances-survey},
but defining such rules for complex scenarios is infeasible. 

Recent learning-based methods capture general affordance features to enable interaction in simulation~\cite{where2act, where2explore, xu2022umpnet, environment-aware-affordances},
yet they often struggle to generalize to real-world environments due to the sim-to-real gap~\cite{Tobin2017DomainRF}.
Other approaches learn from real-world human interaction videos~\cite{hands-as-probes, visual-robotics-bridge, ha-envaff_ego, ha-hand_mot_inter_hot, affordance-diffusion,srirama2024hrp},
but face visual ambiguities or embodiment mismatches.
Moreover, they often assume different properties to be known in advance, such as actions~\cite{where2act, where2explore, environment-aware-affordances}, objects~\cite{where2act, where2explore, environment-aware-affordances, xu2022umpnet, ju2024robo},
movable parts~\cite{where2act, where2explore, environment-aware-affordances}, or segmentation~\cite{where2act, where2explore, environment-aware-affordances, xu2022umpnet, ju2024robo}.

Our approach avoids these assumptions by factorizing affordances into two simpler properties: graspability and movability. While we also use a model trained on human data~\cite{hands-as-probes} to determine movability, we avoid embodiment mismatch by coupling with a robot-specific graspability model~\cite{contact-graspnet}. This coupling also aids in disambiguating affordance cues through complementarity of the two estimation processes.

\subsection{Coupled Recursive Estimation}\label{sec:aicon}

Coupled recursive estimation encodes information about perceptual sub-aspects and their relationships, enabling robust interpretation of sensory signals through Bayesian inference. This approach has been applied to the estimation of kinematic structure and other dynamic properties of objects~\cite{roberto-martin-martin} as well as the overall object segmentation~\cite{vito_motion}. Its semi-decomposed structure allows integration of additional sensor modalities through simple composition, such as proprioception~\cite{roberto-martin-martin-crossmodal}, audio signals~\cite{baum-audio}, or hand poses of surrounding humans~\cite{pfisterer2025helpinghands}. 

Unlike traditional distributed state estimation methods~\cite{distributed_kalman_filter_consensus, LC_Distributed_Particle_Filters}, coupled recursive estimation enforces structured interactions between estimators, akin to factor graphs~\cite{dellaert2017factor} but more specific. This structure constrains system dynamics, enabling complex action generation with simple gradient descent~\cite{mengers2025noplan}, and shares similarities with biological perceptual processes, motivating its use as a model of human vision~\cite{mengers2025robotics,battaje_aicon}.

We adopt this framework for affordance estimation by coupling particle filters that operate on different signals and representations, graspability and movability. While prior work has explored coupling filters over the same state variables~\cite{jacob2016coupling}, our approach coordinates multiple, distinct estimators via cross-modal fusion, as we explain in detail in the following.

\begin{figure}  
    \vspace*{-0.28cm}
    \centering
    \begin{minipage}{\columnwidth}  
        \begin{algorithm}[H]  
        \caption{Coupled Particle Filters}
        \label{alg:algo}
        \begin{algorithmic}[1]
            \STATE \textbf{Initialize:} Two sets of particles $\{x^i_1\}_{i=1}^N$ and $\{x^i_2\}_{i=1}^N$.
            \FOR{each time step $t$}
                \FOR{each Filter $f \in \{1,2\}$}
                    \STATE \textbf{Prediction Step:} Propagate particles according to optical flow.
                    \STATE \textbf{Injection Step:} Inject new particles in low-particle-density areas.
                    \STATE \textbf{Update Step:} Weight particles using neural network measurements.
                \ENDFOR
                \STATE \textbf{Cross-Modal Fusion:} Weigh particles based on the other belief's density.
                \FOR{each Filter $f \in \{1,2\}$}
                    \STATE \textbf{Resample:} Draw particles from belief according to normalized weights.
                \ENDFOR
            \ENDFOR
        \end{algorithmic}
        \end{algorithm}
    \end{minipage}
\end{figure}

\begin{figure*}[t]
    \vspace*{0.17cm}
    \centering
    \begin{minipage}{\textwidth} 
        \centering
        \includegraphics[width=\textwidth]{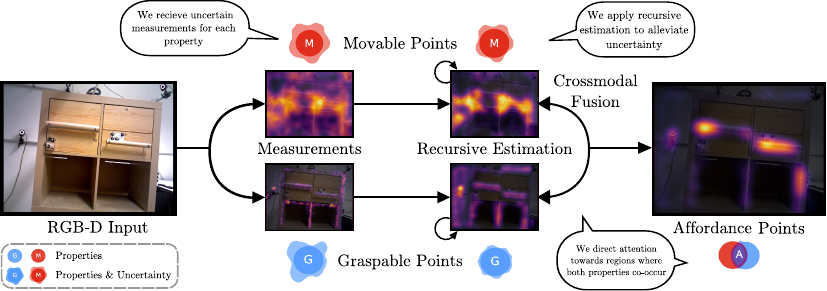}  
        \caption{Our approach estimates affordances as a combination of the graspable and movable regions of our environment. Since our measurement sources provide only uncertain measurements for these properties, we recursively estimate a belief for them and apply additional priors to resolve ambiguities. This aggregates a strong belief in regions with high measurements for their individual property. However, affordances need to be both graspable and movable. Therefore, we exchange information between our estimators, focusing each estimator's attention on relevant parts of the scene that satisfy both properties, i.e., affordances.}
        \label{fig:cross_column_example}
    \end{minipage}
\end{figure*}

\section{Coupled Particle Filters for Affordances}\label{sec:hpf}

We factorize affordances into graspability and movability, estimate these separately, and couple them to obtain an embodiment-appropriate and disambiguated belief over affordances. Both estimation problems are multi-modal, motivating the use of particle filters, where we incorporate additional priors to address property-specific ambiguities. To couple them, we implement a cross-modal information exchange that directs each estimator toward mutually consistent affordances. First, we describe the template for both particle filters in \Cref{sec:standard_procedure}, then their instantiation for movability in \Cref{sec:specific_move}, for graspability in \Cref{sec:specific_grasp}, and finally their fusion in \Cref{sec:crossmodal_fusion}.

\subsection{Shared Particle Filter Procedure}\label{sec:standard_procedure}
Estimating either graspability or movability requires reasoning over noisy, ambiguous observations across many candidate regions.
We measure each property with a learning-based model from the literature, then refine the estimate with a particle filter that enforces additional constraints. 
Here, we describe the shared procedure before detailing the property-specific instantiations in the next two sections.

Each particle represents a 3D position $x \in \mathbb{R}^3$ corresponding to a graspable or movable region. 
Rather than initializing particles uniformly, we draw them from the first measurement (\Cref{alg:algo}, L.~1) to
concentrate particles near likely states and improve initial convergence. 
We then recursively refine beliefs through prediction and update steps (\Cref{alg:algo}, L.~2–7).

During prediction, particles are propagated according to optical flow with added noise for diversity to account for motion in the scene. 
Optical flow measurements are often ambiguous in cluttered scenes, so we track particle depth and reject physically inconsistent motion. 
To recover from local minima and detect new objects, we include an injection step (\Cref{alg:algo}, L.~4), which samples new particles in low-density regions, where new objects are most likely.

After prediction and injection, each filter updates particles using its own measurement source and then incorporates information from the coupled belief, before resampling the particle set.

\subsection{Specific Instantiation for Movability}\label{sec:specific_move}

Movability is measured using Hands-as-Probes~\cite{hands-as-probes}, which predicts a heatmap of movable regions from RGB images of human interactions.
To weight particles according to this measurement, we project their 3D positions into the heatmap and assign the corresponding value as weight $\tilde{w}_\mathrm{movability}$.

Because Hands-as-Probes~\cite{hands-as-probes}  was trained on human interactions, its predictions do not directly transfer to robots.
To address this mismatch, we restrict the movable estimator to regions that a robot can interact with by incorporating information about robotic grasps through its coupling with the graspability estimator.
This adjustment focuses the movable estimator on affordances for robots, resolving further noise and ambiguities.

\subsection{Specific Instantiation for Graspability}\label{sec:specific_grasp}

Graspability requires modeling not only location but also grasp pose. Each particle thus represents not just a 3D position $x \in \mathbb{R}^3$ but also an associated gripper pose $\nu \in \mathbb{SE}(3)$.

As a measurement source, we use GraspNet~\cite{contact-graspnet}, which predicts candidate grasps with contact points and success likelihoods from a scene point cloud. To thus ensure that the tracked particle set contains graspable points with appropriate grasps, we determine particle weights by comparing both grasp likelihood and pose similarity to nearby GraspNet predictions. 

Let $\mathbb{Z}_t={(y_j\in\mathbb{R}^3,\upsilon_j\in\mathbb{SE}(3),p_j\in[0,1])}$ be the set of candidate grasps predicted by GraspNet at time $t$, where $y_j$ is a 3D grasp position, $\upsilon_j$ is a gripper pose, and $p_j$ is the associated predicted grasp success likelihood. To weight a particle with position $x \in \mathbb{R}^3$ and gripper pose $\nu \in \mathbb{SE}(3)$, we first ignore gripper orientation and only consider spatial proximity, encoded through a Gaussian kernel for distance $K(x,y)$, and predicted grasp success $p_j$ to determine a grasp success likelihood:
\begin{equation}
	    v = \max_{(y_j, \upsilon_j, p_j)\in \mathbb{Z}_t}{p_j\;K(x,y_j)}.
\end{equation}
Then we furthermore consider the similarity of grasps, using cosine similarity $S(\nu,\upsilon_j)$ of gripper approach vectors, while considering their overall likelihood $p_j$ and proximity: \begin{equation}\label{eq:graspsim}
	u = \max_{(y_j,\upsilon_j,p_j)\in \mathbb{Z}_t}{p_j\;K(x,y_j)\; S(\nu,\upsilon_j)}.
\end{equation}

The final weight is then the product of these two similarity measures, i.e., $\tilde{w}_\mathrm{graspability}=u\cdot v$. Multiplying both ensures that particles are weighted highly only if they are both near a high-confidence grasp and have a compatible orientation. To further reduce false positives at geometric ambiguities (e.g., depth discontinuities), we detect such regions in the depth image and down-weight particles accordingly.

These tailored filters allow each estimator to handle its modality-specific ambiguities and adapt to dynamic environments through targeted particle injections. However, each focuses only on its own property, not on the joint affordances of interest. To achieve this coordination, we introduce coupling between the filters, described next. 

\subsection{Coupling Estimators for Cross-Modal Fusion}\label{sec:crossmodal_fusion}
To align both estimators on manipulable regions consistent with the robot's embodiment, we couple them through cross-modal fusion. This biases resampling toward particles supported by both beliefs, effectively shifting each estimator's attention to mutually consistent regions.

For this coupling, we want to weight each particle depending on its likelihood according to \emph{both} beliefs. For each set, we first determine its measurement-specific weight $\tilde{w}$ in its respective estimator. Then we determine its likelihood given the particle set of the other estimator. For each estimator with particle set $\mathbb{X}$, we determine the cross modal weight for each particle by comparing it to the density of the other estimator's particle set $\mathbb{Y}$ (\Cref{alg:algo}, L.~8). To do so, we represent both beliefs in 3D and estimate densities using a Gaussian kernel with $\sigma=0.05$ to the cross-modal weight $\hat{w}$ for particle $x\in\mathbb{X}$:
\begin{equation}\label{eq:kernel}
	\hat{w} = \sum_{y \in \mathbb{Y}} \exp\left(-\frac{\|x-y\|^2}{2 \sigma^2}\right).
\end{equation}

The final particle weight in each estimator then combines the original measurement-based weight $\tilde{w}$ with the cross-modal weight $\hat{w}$ for each particle:
\begin{equation}
	w = f_u(\tilde{w})\, f_c(\hat{w}),
	\label{eq:finalupdate}
\end{equation}
where $f_u$ and $f_c$ clip values to predefined ranges tuned on a single scene. For graspability, we clip $\tilde{w}$ to $(0.2, 0.25)$ and $\hat{w}$ to $(0.3, 0.6)$. For movability, we clip $\tilde{w}$ to $(0.2, 0.3)$ and $\hat{w}$ to $(0.45, 0.55)$.

We assign broader ranges to the cross-modal weights, giving them higher influence since beliefs are typically more reliable than raw measurements. This ensures particles are more likely resampled if supported by both estimators, enforcing mutual agreement. The result is a robust joint belief over affordances that integrates complementary priors and, in the case of graspability, already encodes executable grasps for direct interaction. 

\section{Experiments}
To evaluate our approach, we compare against three recent learning-based methods on a real-world dataset. First, we describe the experimental setup, including how we extend an existing dataset with affordance annotations, (\Cref{sec:setup}). Next, we demonstrate that fusing graspable and movable information improves performance over recent approaches (\Cref{sec:exp_performance}), while also providing robustness under challenging conditions (\Cref{sec:exp_robustness}). Lastly, we show how this effectiveness stems not from our measurement sources but the coupling of estimators (\Cref{sec:ablation1}), and how cross-modal coupling is beneficial compared to other fusion strategies (\Cref{sec:ablation2}).

\subsection{Experimental Setup}\label{sec:setup}

To evaluate the robustness of affordance estimation in real-world scenarios, we need a dataset for such affordances given real sensor data. To our knowledge, existing datasets lack explicit manipulable region annotations~\cite{damen2020epic,grauman2022ego4d,visual-robotics-bridge}, motivating our extension of the RBO dataset~\cite{rbo-dataset} (\Cref{sec:dataset}). On this dataset, we compare against three recent works on affordance estimation (\Cref{sec:baseline}), using appropriate metrics (\Cref{sec:metrics}).

\subsubsection{Dataset}\label{sec:dataset}

We evaluate our approach using the RBO dataset of articulated objects and interactions~\cite{rbo-dataset}, which contains RGB-D videos of interactions with various articulated objects in challenging environmental conditions. Additionally, it includes ground-truth models and poses for the articulated objects in the scene, enabling precise evaluation. To assess performance, we annotate the manipulable parts of these object models with interaction volumes (3D regions indicating where a robot can interact) and project them into the sequences.

To evaluate performance and robustness against adverse conditions, we further split the dataset into three parts: well-lit, dark, and cluttered environments. For cluttered environments, we only show qualitative examples, since the dataset does not include object models for the distractor objects.

\subsubsection{Learning-Based Approaches}\label{sec:baseline}
We compare against three recent learning-based affordance estimation techniques to show both the performance and robustness of our approach. Where2Act~\cite{where2act} predicts grasps and success likelihood from point cloud data and an action primitive, while Hands-as-Probes~\cite{hands-as-probes} and a vision transformer trained with HRP~\cite{srirama2024hrp} predict heatmaps for possible interactions given human interaction data. Since the vision transformer predicts 12 heatmaps, we average their values to a single heatmap.

\subsubsection{Metrics}\label{sec:metrics}

To quantify the performance of the different approaches, we compute the mean precision across all sequences for each object. For this, we sample $200$ affordance points from the prediction of each approach to ensure comparable coverage across methods. We then count the proportion of samples within $2$cm of the nearest manipulable region. This tolerance accounts for inaccuracies between shape models and transformations. For objects with multiple manipulable parts, we additionally analyze the distribution across these regions to show if the models detect all relevant affordances.

\subsection{Coupling Reduces Uncertainty and Improves Precision}\label{sec:exp_performance}

\begin{figure}[t]
    \vspace{0.2cm}
    \centering
    \includegraphics[width=\linewidth]{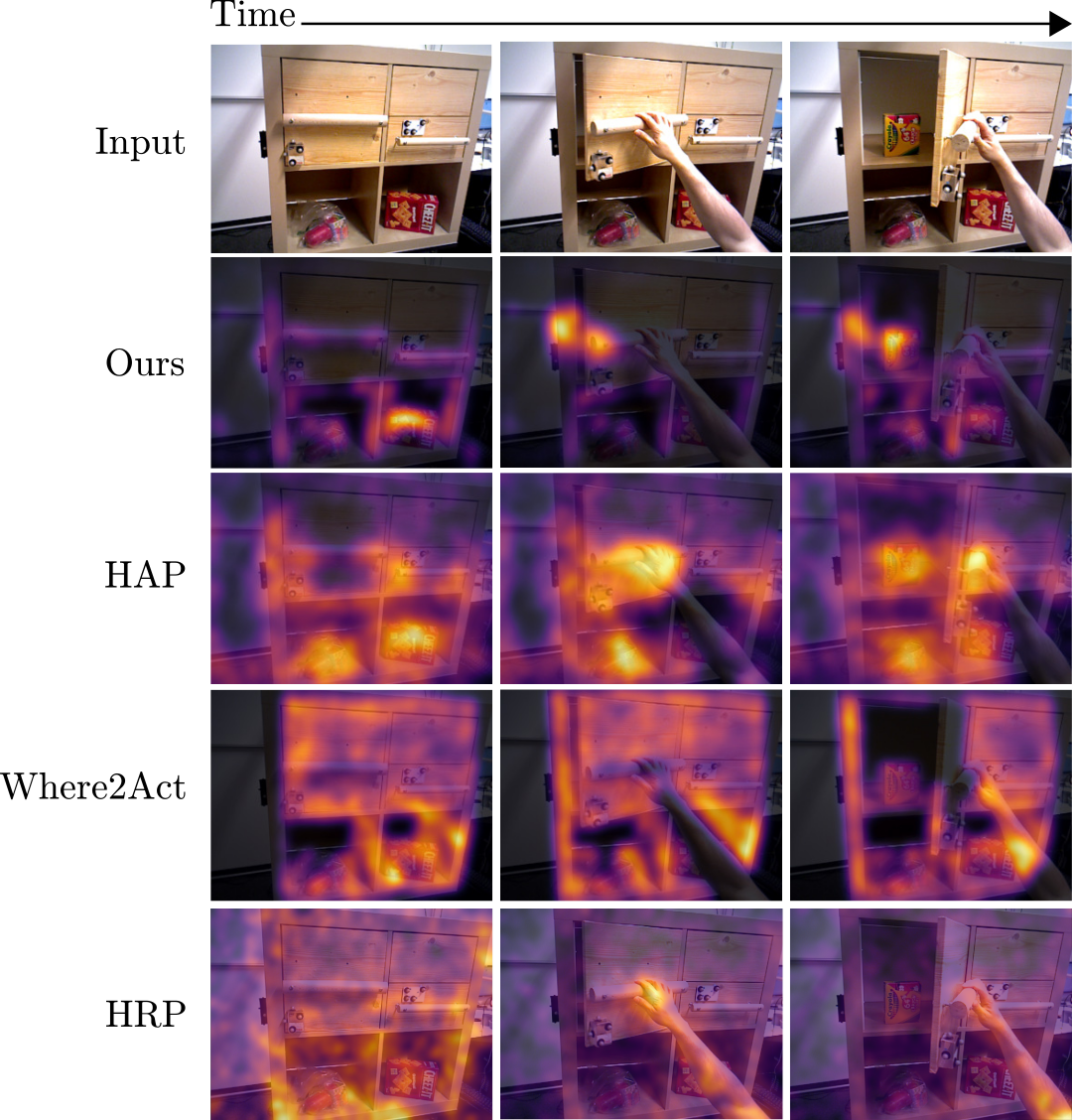}
    \caption{Our method can handle dynamic and complex scenes. While the non-recursive learning-based approaches detect new affordances directly, we rely on our injection step to introduce new affordances, such as the object inside the shelf, into our belief. Our estimates are more precise, focusing on manipulable affordances.}
    \label{fig:clutter_in_ikea}
\end{figure}

Our method couples estimators for graspable and movable areas and aligns their attention on regions where these properties co-occur, i.e., affordances. 
In these regions, the estimators share complementary information, resolving ambiguities that single modalities cannot handle.  

We demonstrate the improved performance of our approach by comparing it against three recent affordance estimators, Where2Act~\cite{where2act}, Hands-as-Probes~\cite{hands-as-probes}, and HRP~\cite{srirama2024hrp}, on well-lit and uncluttered scenes from the RBO dataset~\cite{rbo-dataset}.  
As shown in~\Cref{fig:scores}, our approach achieves the highest mean precision across all objects, outperforming Where2Act~\cite{where2act} by 308\%, Hands-as-Probes~\cite{hands-as-probes}  by 245\%, and the HRP model~\cite{srirama2024hrp} by 257\%.  

All three learning-based approaches struggle with out-of-distribution data and modality-specific ambiguities in their input representation: Where2Act~\cite{where2act} with geometric and Hands-as-Probes~\cite{hands-as-probes} and the HRP model~\cite{srirama2024hrp} with visual ambiguities (\Cref{fig:clutter_in_ikea}).  
Our method addresses these challenges by filtering noisy measurements, fusing complementary modalities, and incorporating priors. Grasps encode geometric cues that disambiguate flat or textured regions, while movability estimates and priors suppress grasps in non-movable areas.  

When the estimators align, they exchange refined information, yielding localized and reliable affordance predictions. Next, we evaluate how this robustness extends to more challenging environments.  

\subsection{Coupling Improves Robustness in Adverse Conditions}\label{sec:exp_robustness}
We further evaluate robustness under adverse conditions, showing that our method remains reliable:
\begin{figure}[!t]
    \centering
    \includegraphics[width=0.97\linewidth]{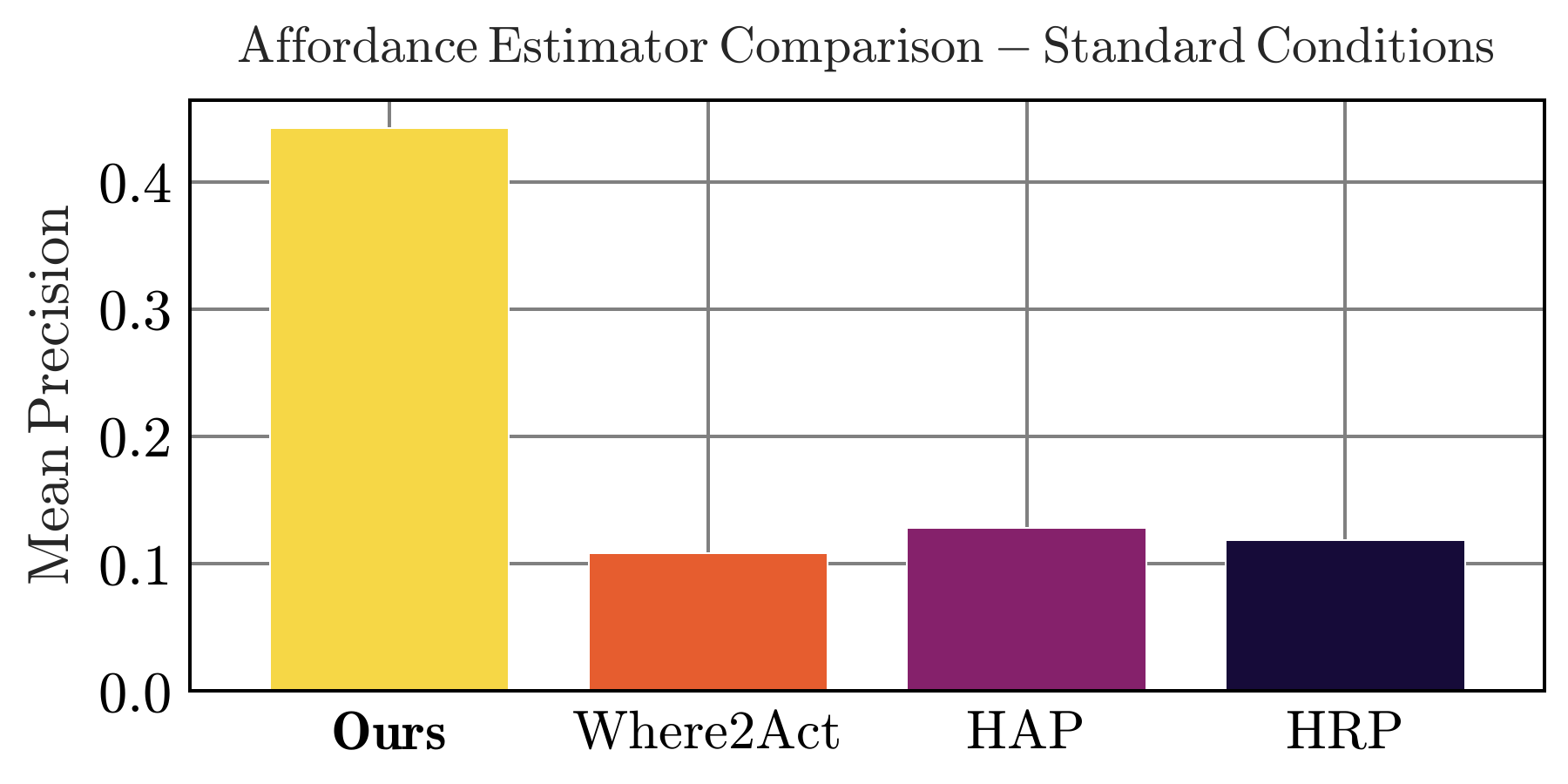}
    \includegraphics[width=0.97\linewidth]{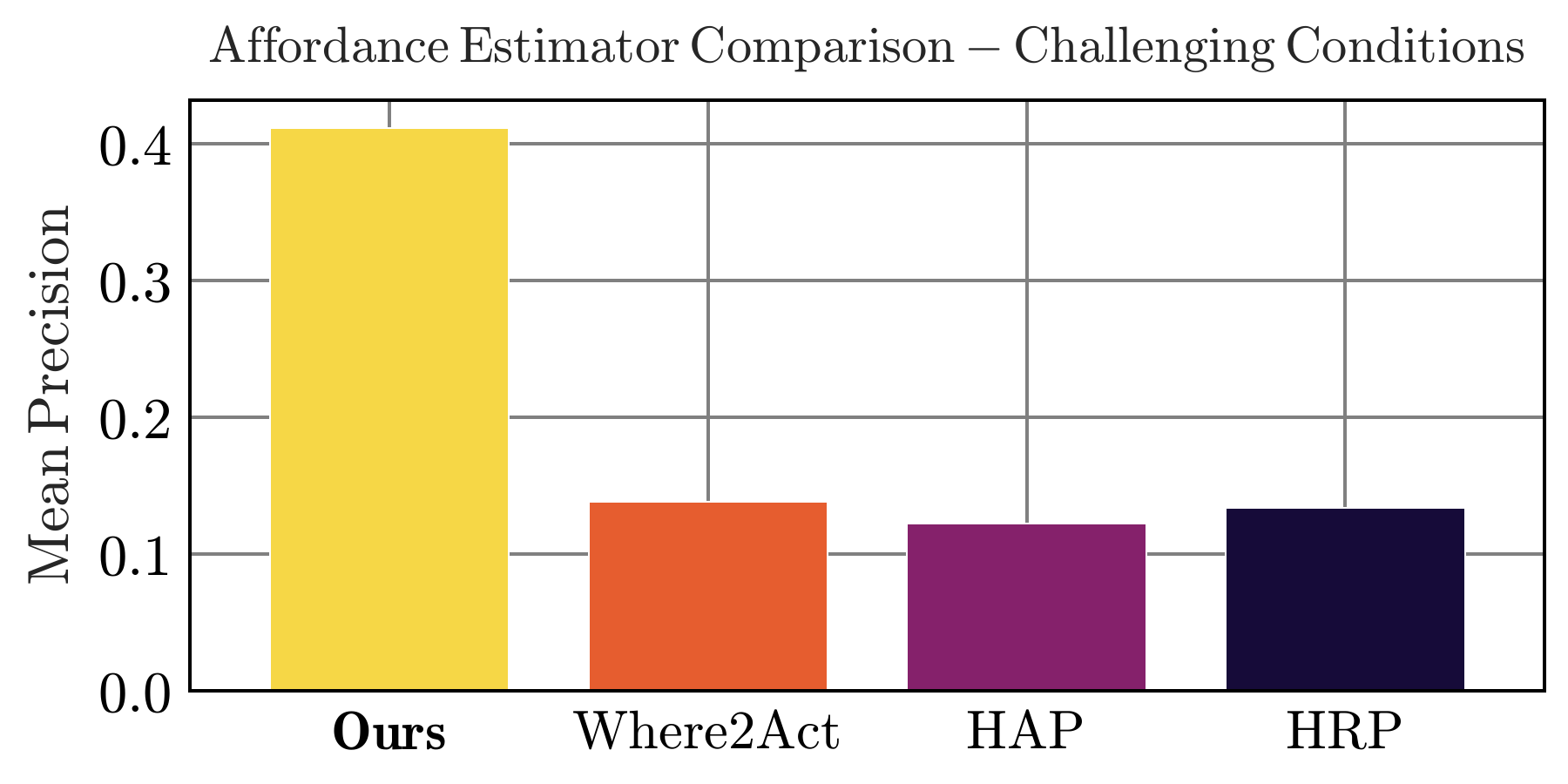}
    \caption{Our approach outperforms recent learning-based approaches in both standard, i.e., well-lit and clutter-free (top), and more challenging, i.e., dark or cluttered (bottom), settings. The monolithic learned models struggle with real-world ambiguities, while coupling resolves them by aligning estimator attention and filtering noise.}\label{fig:scores}
\end{figure}
\subsubsection{Dark Environments}

Dark environments introduce visual ambiguities that affect the appearance-based Hands-as-Probes~\cite{hands-as-probes} , which also acts as our movable measurement source. While appearance-based methods struggle under these conditions, our coupled system resolves ambiguities by leveraging grasp information.  
As a result, we outperform Hands-as-Probes~\cite{hands-as-probes}  by 235\% and the HRP model~\cite{srirama2024hrp} by 207\% in dark environments. Because grasp estimates remain unaffected by lighting, we also outperform Where2Act~\cite{where2act} by 197\% (\Cref{fig:scores},~bottom).  

\begin{figure}[t]
    \vspace*{0.15cm}
    \centering
    \includegraphics[width=\linewidth]{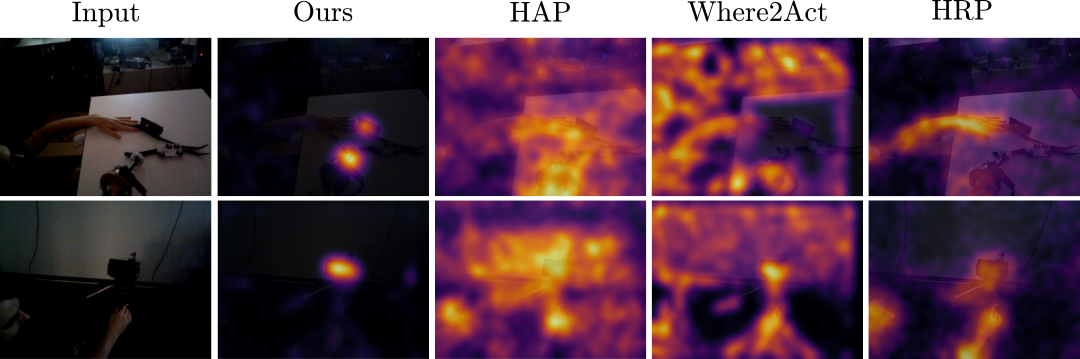}
    \caption{Our affordance estimates remain precise in dark environments. Hands-as-Probes~\cite{hands-as-probes} and HRP~\cite{srirama2024hrp} suffer from visual ambiguities, and Where2Act~\cite{where2act} struggles with object complexity, while our coupled system yields reliable affordances.}
    \label{fig:quali_dark}
\end{figure}

By fusing modalities, our method produces precise and robust affordances despite reduced visual information (\Cref{fig:quali_dark}).  
This robustness to visual ambiguity motivates evaluating our method in cluttered environments, which introduce additional geometric and visual challenges.

\subsubsection{Cluttered Environments}

\begin{figure}[t]
    \vspace*{0.15cm}
	\centering
	\includegraphics[width=\linewidth]{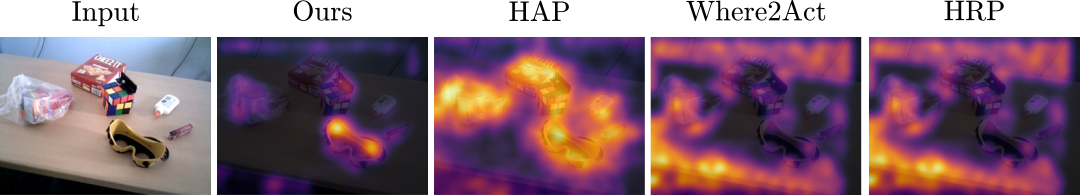}
	\caption{Due to the coupling, our affordance estimates remain focused on manipulable object regions even in clutter. Other approaches are less precise, spreading across distractor surfaces such as the table edge.}
	\label{fig:clutter_on_table}
\end{figure}

Clutter introduces occlusions and texture complexity that hinder both geometry- and appearance-based estimators.  
Qualitative results (\Cref{fig:clutter_on_table,fig:clutter_in_ikea}) show our estimates remain localized on manipulable parts of the target object, while the predictions of recent learning-based approaches often spill into irrelevant regions.  

Our coupled estimators drive attention toward relevant affordances, filtering out spurious grasps such as edges of tables.  
The cluttered IKEA scene (\Cref{fig:clutter_in_ikea}) highlights robustness in dynamic, complex environments and underscores the method’s potential for real-world robotic interaction. Next, we will show in two ablations that this performance and robustness does not stem from our measurement sources or filtering by themselves, but our coupling approach.

\subsection{Ablation: Our Cross-Modal Coupling Aligns Estimators}\label{sec:ablation1}

\begin{figure}[t]
    \centering
    \includegraphics[width=0.97\linewidth]{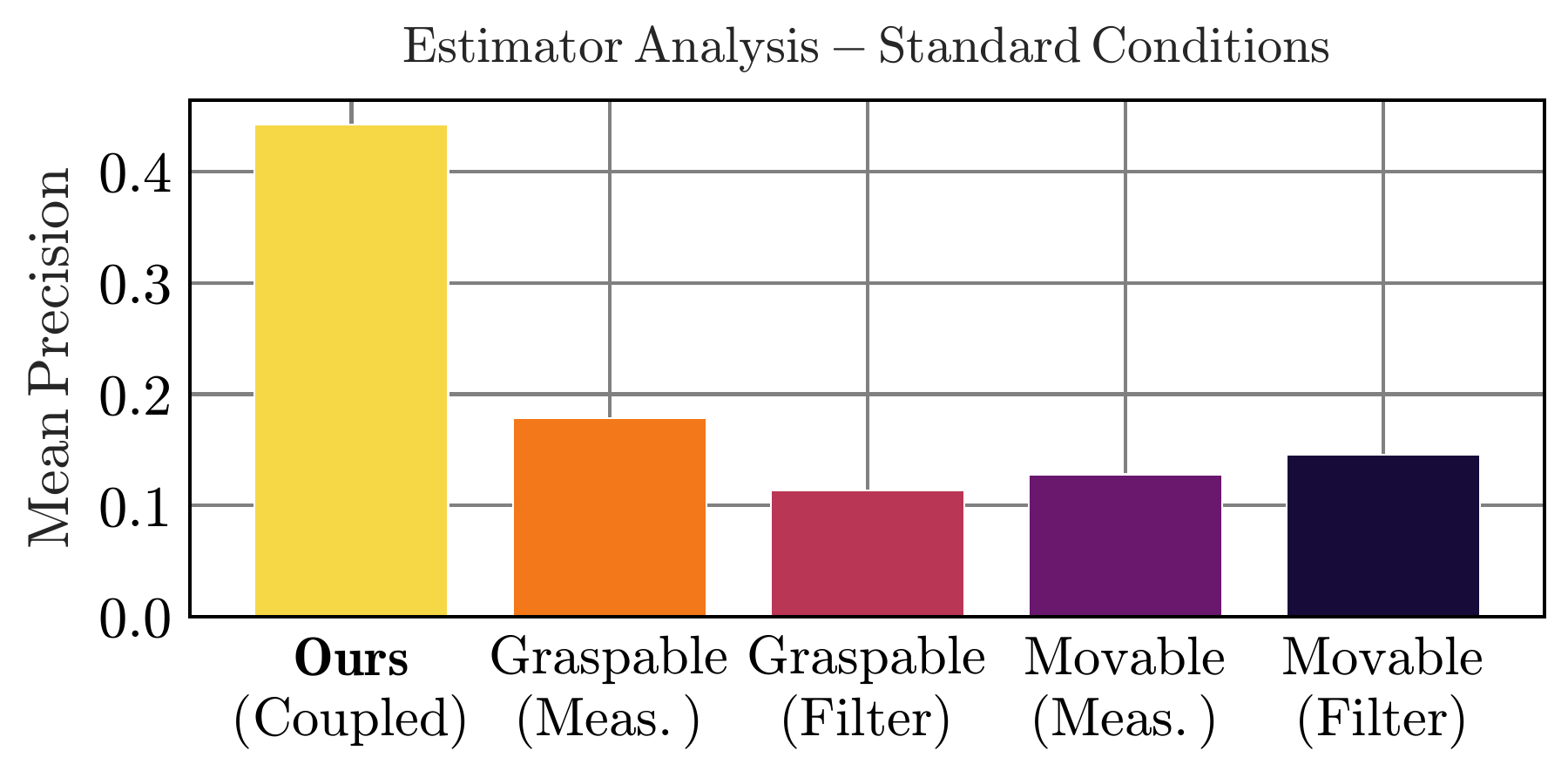}
    \includegraphics[width=0.97\linewidth]{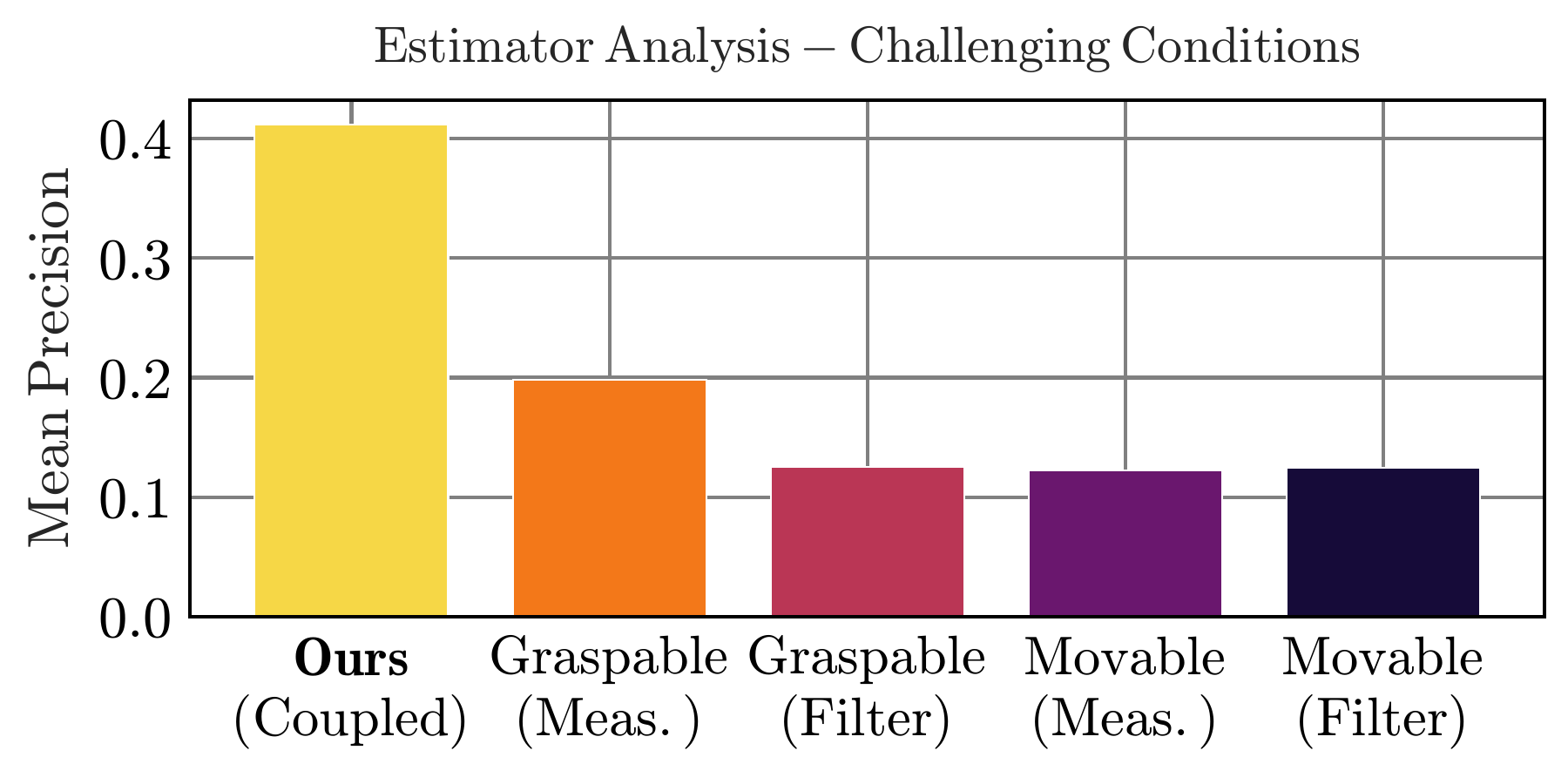}
    \caption{Our cross-modal fusion aligns the estimators' attention on regions that satisfy both properties, that is the affordances. The individual estimators suffer from their respective ambiguities, but by sharing information they can resolve these, focus on relevant regions and estimate robust affordances even under adverse conditions.}
    \label{fig:filtering}
\end{figure}

Our method aligns the estimators for graspable and movable regions on affordances. We show this by analyzing our individual measurement sources and recursive estimators without coupling.
The individual measurement sources struggle to estimate affordances as they only account for one aspect of robotic manipulation (\Cref{fig:filtering}).
Consequently, filtering the grasp prediction results in a performance drop because highly graspable areas, such as table edges, are not interactable. Movability, on the other hand, is a stronger indicator for interaction, and filtering improves performance by 13\% under good lighting conditions. 
However, both measurement sources are susceptible to perceptual ambiguities, which we cannot resolve using data from one modality alone.

By enabling the estimators to inform each other, they can resolve these ambiguities and align their attention on manipulable regions. As a result, our coordinated attention estimates precise and robust affordances even under adverse conditions(\Cref{fig:filtering}). Next, we will show that this cannot be achieved with other fusion methods.

\subsection{Ablation: Coupling of Particle Filters Outperforms Standard Sensor Fusion}\label{sec:ablation2}

\begin{figure}[t]
    \centering
    \includegraphics[width=0.97\linewidth]{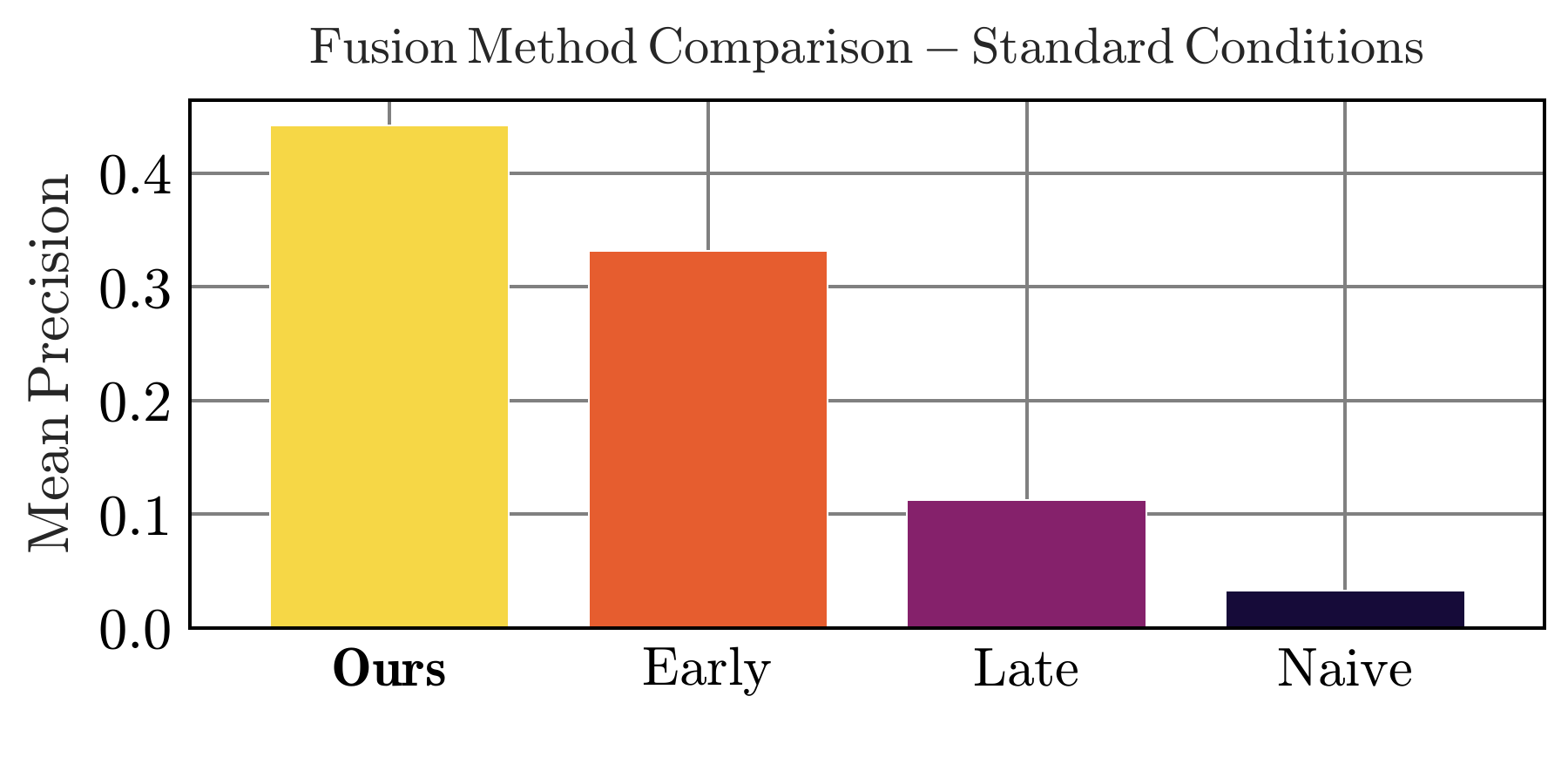}
    \includegraphics[width=0.97\linewidth]{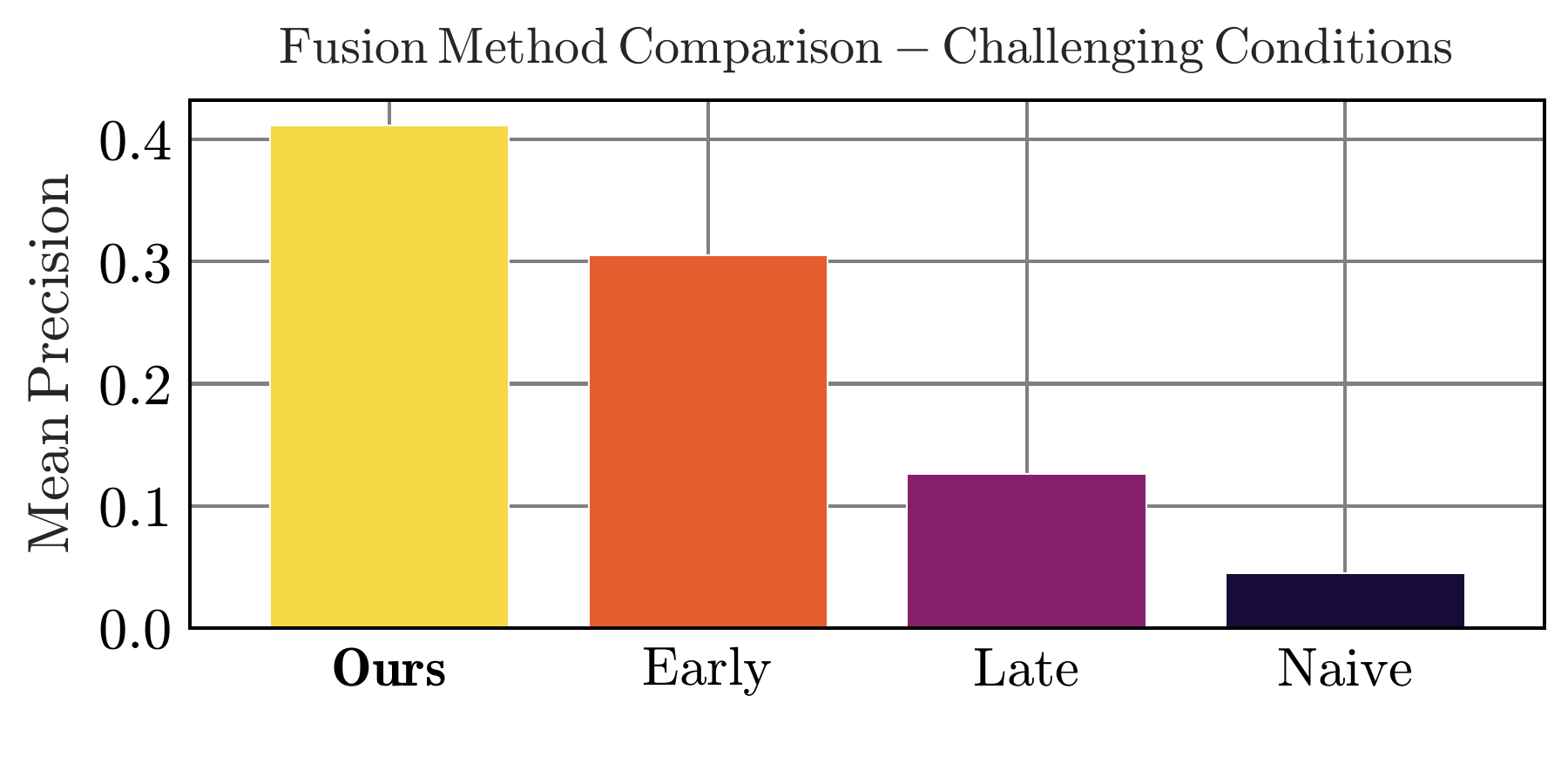}
    \caption{Coupling particle filters with our cross-modal fusion outperforms other sensor fusion schemes, in particular, naive and late fusion. Early fusion combines signals before recursive estimation, which leads to good estimates, but does not leverage the shared structure of the two properties to further disambiguate the signals.}
     \label{fig:sensorfusion}
\end{figure}

In contrast to standard sensor fusion, our approach enables the two estimators to influence and inform each other through cross-modal coupling.
To demonstrate the benefits of this method, we compare it to standard sensor fusion approaches~(\Cref{fig:sensorfusion}).

Our approach outperforms all other sensor fusing methods, as it accounts for uncertainty and the relationship between the properties.
Naive fusion directly combines the estimates for graspability and movability, but this does not account for the uncertainty in both measurement sources.
Early and late fusion, on the other hand, alleviate measurement uncertainty through recursive estimation but do not model the relationship between the properties.
Late fusion processes the signals independently and thus loses the shared structure. By combining the signals before we apply recursive estimation, we keep the shared structure in the signal but do not explicitly leverage it.

Our method utilizes the relationship between the properties in the interconnection between estimators and addresses uncertainty, and can therefore estimate more robust and precise affordances.

\subsection{Real World Experiments}\label{sec:real-world}

We conduct real-world experiments to evaluate whether affordances that score highly on benchmark datasets also correspond to actionable grasps in physical scenes. We again compare against Hands-as-Probes~\cite{hands-as-probes} and HRP~\cite{srirama2024hrp}, and exclude Where2Act~\cite{where2act} because obtaining the required pointcloud segmentation for real-world scenes is a challenging perception problem by itself. For each trial, we compute the highest-scoring point in the heatmap or the densest region in our belief after 10 iterations and use this as our affordance estimate. If a robot can grasp and move the estimated affordance, we count the trial as a success.

On cluttered tabletops, our method achieves an 80\% success rate, outperforming Hands-as-Probes and HRP, which each achieve 40\%. In the more challenging cluttered IKEA shelf scenes, our method achieves a 60\% success rate, compared to 20\% for Hands-as-Probes and 40\% for HRP. These results represent up to a threefold improvement over prior methods in difficult scenes. 

The appearance-based models often struggle with visual ambiguities and predict affordances at non-manipulable regions. In contrast, our approach resolves many of these ambiguities by incorporating grasp information into the affordance estimation process. Failures in the shelf scenes typically occur due to systemic errors in both measurement sources.

Across all 10 trials, our method achieves a 70\% success rate, outperforming the learned baselines and demonstrating the robustness of our coupled estimators in challenging real-world environments.

\begin{figure}[t]
    \vspace{0.2cm}
    \centering
    \includegraphics[width=\linewidth, right]{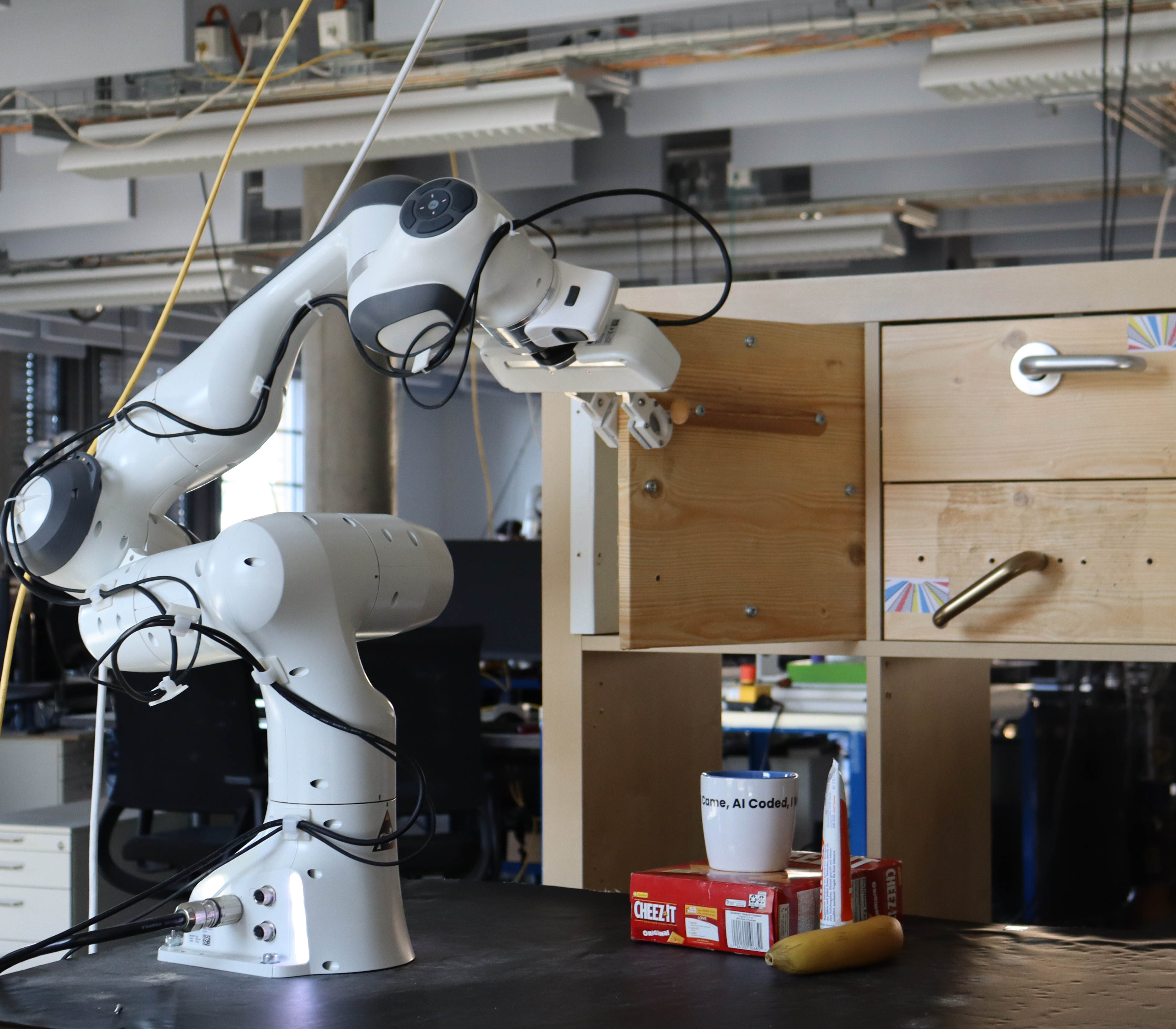}\\[\baselineskip]
    \includegraphics[width=\linewidth, right]{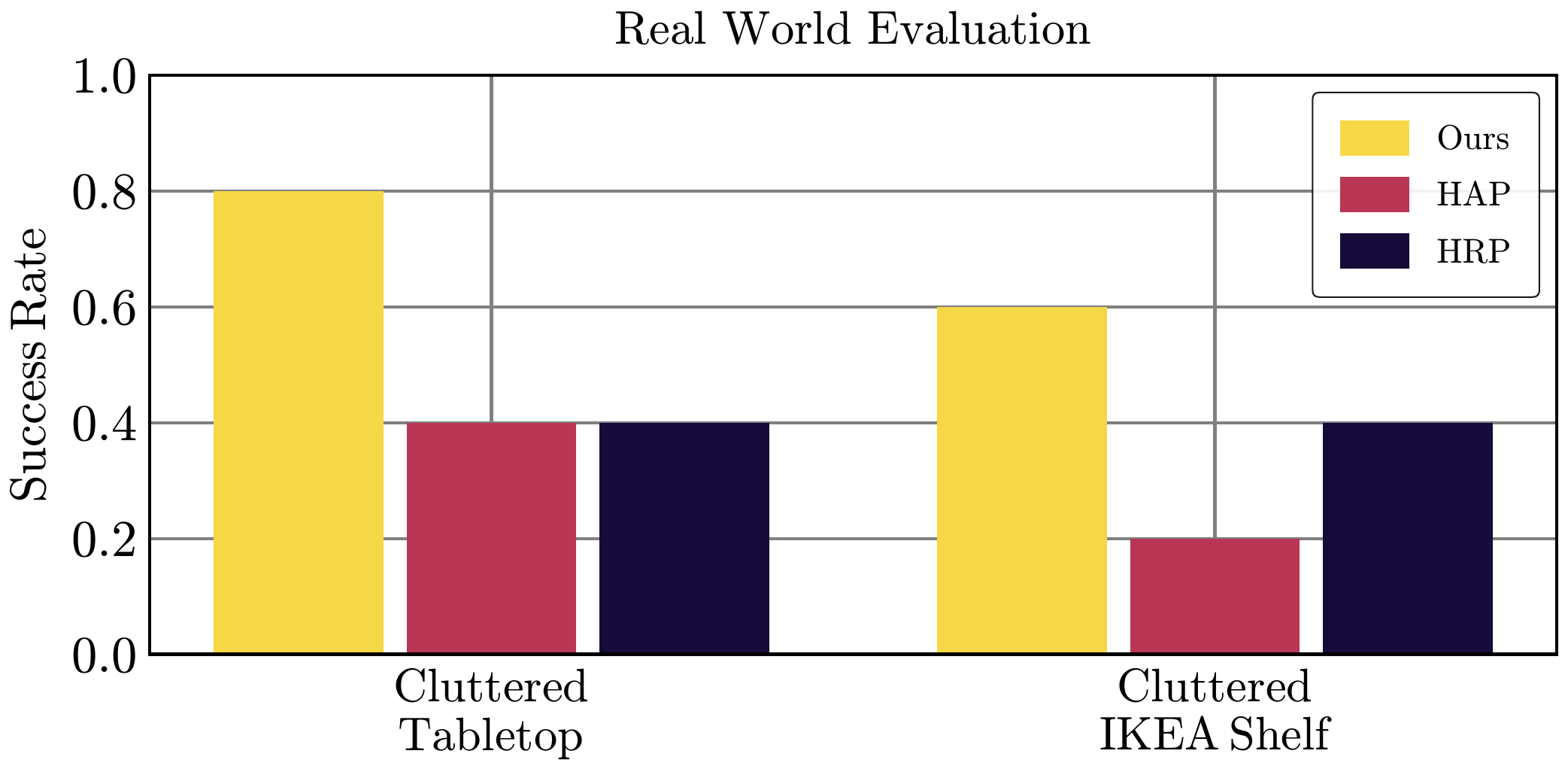}
    \caption{Our approach estimates robust affordances that enable robots to interact with their environment. We evaluate this in real-world trials on cluttered tabletop and shelf scenes with four objects (cup, banana, toothpaste, and Cheez-it box). Our estimated affordances correspond to precise and actionable grasps, even in cluttered environments.}
    \label{fig:realworld}
\end{figure}

\section{Limitations \& Discussion}

Our approach proved remarkably robust throughout our experiments, but is still limited by the accuracy and reliability of its measurement models. In particular, systematic errors that occur in both graspability and movability estimators cannot be resolved through coupling alone. This could be addressed by introducing additional complementary estimators, either based on additional visual properties or based on other sensors, such as tactile or audio cues. Alternatively, we could make the system interact with its environment to resolve ambiguities by gathering information from different viewing directions or eliminate non-manipulable regions by interacting with the estimated affordances.

A second limitation is computational cost. The coupled particle filters themselves add little overhead, but the measurement models rely on large neural networks, particularly for movability prediction with Hands-as-Probes~\cite{hands-as-probes}. Replacing these with faster models would allow real-time affordance estimation without altering the overall pipeline.  

Despite these limitations, our approach has several strengths: it substantially improves robustness over raw measurements, is lightweight once measurements are available, and as a modular system it can easily be extended with additional estimators. Parameter tuning proved straightforward, optimized manually on only two sequences, yet generalized across the dataset thanks to the strong inductive bias of the coupled estimation framework and Bayesian inference. This suggests a promising route to enhance the generalizability of modern learning-based perception methods beyond their training assumptions, especially as an increasing number of estimators is incorporated.

\section{Conclusion}
We propose a method to estimate actionable affordances in real-world environments by identifying regions that are both graspable and movable. To address uncertainty of our learning-based measurement models, we combine recursive estimation with coupling between graspability and movability estimators. This alignment allows the estimators to exchange complementary information, resolve ambiguities, and yield robust affordance predictions fitting the robotic embodiment.

In experiments on the RBO dataset~\cite{rbo-dataset} of articulated objects, our method outperforms recent learning-based affordance estimators by large margins and demonstrates robustness under adverse conditions such as low light and clutter. The resulting affordances are precise, localized, and probabilistic, supporting reasoning about uncertainty. Together, these properties make our approach well suited for robotic exploration and manipulation, as seen in our real-world experiments, and point toward scalable, modular perception systems built from coupled estimators, capable of reasoning over multiple complementary sensory cues and models.

\newpage
\balance


\bibliographystyle{IEEEtran-nourl}
\bibliography{bibliography}

@article{contact-graspnet,
  author={Sundermeyer, Martin and Mousavian, Arsalan and Triebel, Rudolph and Fox, Dieter},
  journal={IEEE International Conference on Robotics and Automation (ICRA)}, 
  title={{Contact-GraspNet}: Efficient {6-DoF} Grasp Generation in Cluttered Scenes}, 
  year={2021},
  pages={13438-13444},
}

@article{visual-affordances-survey,
author = {Hassanin, Mohammed and Khan, Salman and Tahtali, Murat},
title = {Visual Affordance and Function Understanding: A Survey},
year = {2021},
volume = {54},
number = {3:47},
journal = {ACM Computing Surveys (CSUR)},
pages = {1-35}
}

@article{affordance-tool-detection,
  author={Myers, Austin and Teo, Ching L. and Fermüller, Cornelia and Aloimonos, Yiannis},
  journal={IEEE International Conference on Robotics and Automation (ICRA)}, 
  title={Affordance detection of tool parts from geometric features}, 
  year={2015},
  pages={1374-1381},
}

@article{visual-robotics-bridge,
  title={Affordances from human videos as a versatile representation for robotics},
  author={Bahl, Shikhar and Mendonca, Russell and Chen, Lili and Jain, Unnat and Pathak, Deepak},
  journal={IEEE/CVF Conference on Computer Vision and Pattern Recognition (CVPR)},
  pages={13778--13790},
  year={2023}
}

@article{roberto-martin-martin,
author = {Roberto Martín-Martín and Oliver Brock},
title ={Coupled recursive estimation for online interactive perception of articulated objects},  
journal = {The International Journal of Robotics Research (IJRR)},
volume = {41},
number = {8},
pages = {741-777},
year = {2022},
}

@article{vito_motion,
  author={Mengers, Vito and Battaje, Aravind and Baum, Manuel and Brock, Oliver},
  journal={IEEE International Conference on Robotics and Automation (ICRA)}, 
  title={Combining Motion and Appearance for Robust Probabilistic Object Segmentation in Real Time}, 
  year={2023},
  pages={683-689},
}

@article {hands-as-probes,
author = { Goyal, Mohit and Modi, Sahil and Goyal, Rishabh and Gupta, Saurabh },
journal = {IEEE/CVF Conference on Computer Vision and Pattern Recognition (CVPR)},
title = {Human Hands as Probes for Interactive Object Understanding },
year = {2022},
pages = {3283-3293},
}

@article{roberto-martin-martin-crossmodal,
  author = {Roberto Martín-Martín and Oliver Brock},
  journal={IEEE/RSJ International Conference on Intelligent Robots and Systems (IROS)}, 
  title={Cross-modal interpretation of multi-modal sensor streams in interactive perception based on coupled recursion}, 
  year={2017},
  pages={3289-3295},
}

@article {affordance-diffusion,
author = {Ye, Yufei and Li, Xueting and Gupta, Abhinav and De Mellon, Shalini and Birchfield, Stan and Song, Jiaming and Tulsiani, Shubham and Liu, Sifei },
journal = {IEEE/CVF Conference on Computer Vision and Pattern Recognition (CVPR)},
title = {Affordance Diffusion: Synthesizing Hand-Object Interactions },
year = {2023},
pages = {22479-22489},
}

@article{where2act,
  author={Mo, Kaichun and Guibas, Leonidas and Mukadam, Mustafa and Gupta, Abhinav and Tulsiani, Shubham},
  journal={IEEE/CVF International Conference on Computer Vision (ICCV)}, 
  title={{Where2Act}: From Pixels to Actions for Articulated {3D} Objects}, 
  year={2021},
  pages={6793-6803},
}

@article{where2explore,
author = {Ning, Chuanruo and Wu, Ruihai and Lu, Haoran and Mo, Kaichun and Dong, Hao},
title = {{Where2Explore}: Few-shot affordance learning for unseen novel categories of articulated objects},
year = {2023},
journal = {International Conference on Neural Information Processing Systems (NeurIPS)},
pages = {203},
}

@article{environment-aware-affordances,
author = {Wu, Ruihai and Cheng, Kai and Shen, Yan and Ning, Chuanruo and Zhan, Guanqi and Dong, Hao},
title = {Learning environment-aware affordance for {3D} articulated object manipulation under occlusions},
year = {2023},
journal = {International Conference on Neural Information Processing Systems (NeurIPS)},
pages = {2664},
numpages = {18},
}

@article{ha-envaff_ego,
  title={{Ego-Topo}: Environment affordances from egocentric video},
  author={Nagarajan, Tushar and Li, Yanghao and Feichtenhofer, Christoph and Grauman, Kristen},
  journal={IEEE/CVF Conference on Computer Vision and Pattern Recognition (CVPR)},
  pages={163--172},
  year={2020}
}

@article{ha-hand_mot_inter_hot,
  title={Joint hand motion and interaction hotspots prediction from egocentric videos},
  author={Liu, Shaowei and Tripathi, Subarna and Majumdar, Somdeb and Wang, Xiaolong},
  journal={IEEE/CVF Conference on Computer Vision and Pattern Recognition (CVPR)},
  pages={3282--3292},
  year={2022}
}

@article{rbo-dataset,
  title={The {RBO} dataset of articulated objects and interactions},
  author={Mart{\'\i}n-Mart{\'\i}n, Roberto and Eppner, Clemens and Brock, Oliver},
  journal={The International Journal of Robotics Research (IJRR)},
  volume={38},
  number={9},
  pages={1013--1019},
  year={2019},
}

@article{baum-audio,
  author={Baum, Manuel and Froessl, Amelie and Battaje, Aravind and Brock, Oliver},
  journal={IEEE International Conference on Robotics and Automation (ICRA)}, 
  title={Estimating the Motion of Drawers From Sound}, 
  year={2023},
  pages={697-703},
}

@article{battaje_aicon,
  title={An information processing pattern from robotics predicts unknown properties of the human visual system},
  author={Battaje, Aravind and Godinez, Angelica and Hanning, Nina M and Rolfs, Martin and Brock, Oliver},
  journal={bioRxiv},
  year={2024},
  number={2024.06.20.599814},
}

@article{xu2022umpnet,
	title={{UMPNet}: Universal manipulation policy network for articulated objects},
	author={Xu, Zhenjia and Zhanpeng, He and Song, Shuran},
	journal={IEEE Robotics and Automation Letters (RAL)},
	year={2022},
	publisher={IEEE}
}

@article{ju2024robo,
  title={Robo-abc: Affordance generalization beyond categories via semantic correspondence for robot manipulation},
  author={Ju, Yuanchen and Hu, Kaizhe and Zhang, Guowei and Zhang, Gu and Jiang, Mingrun and Xu, Huazhe},
  journal={European Conference on Computer Vision (ECCV)},
  pages={222--239},
  year={2024},
  organization={Springer}
}

@book{gibson2014ecological,
  title={The ecological approach to visual perception: classic edition},
  author={Gibson, James J},
  year={2014},
  publisher={Psychology press}
}

@article{mengers2025noplan,
	author={Mengers, Vito and Brock, Oliver},
	journal={IEEE International Conference on Robotics and Automation (ICRA)}, 
	title={No Plan but Everything Under Control: Robustly Solving Sequential Tasks with Dynamically Composed Gradient Descent}, 
	year={2025},
	pages={90-96},
}

@article{pfisterer2025helpinghands,
	author={Pfisterer, Adrian and Li, Xing and Mengers, Vito and Brock, Oliver},
	journal={IEEE International Conference on Robotics and Automation (ICRA)}, 
	title={A Helping (Human) Hand in Kinematic Structure Estimation}, 
	year={2025},
}

@article{mengers2025robotics,
  title={A robotics-inspired scanpath model reveals the importance of uncertainty and semantic object cues for gaze guidance in dynamic scenes},
  author={Mengers, Vito and Roth, Nicolas and Brock, Oliver and Obermayer, Klaus and Rolfs, Martin},
  journal={Journal of Vision},
  volume={25},
  number={2},
  pages={6},
  year={2025},
  publisher={The Association for Research in Vision and Ophthalmology}
}

@ARTICLE{distributed_kalman_filter_consensus,
  author={Das, Subhro and Moura, José M. F.},
  journal={IEEE Transactions on Signal Processing (TSP)}, 
  title={{Consensus+Innovations} Distributed Kalman Filter With Optimized Gains}, 
  year={2017},
  volume={65},
  number={2},
  pages={467-481},
  doi={10.1109/TSP.2016.2617827}}

@ARTICLE{LC_Distributed_Particle_Filters,
  author={Hlinka, Ondrej and Slučiak, Ondrej and Hlawatsch, Franz and Djurić, Petar M. and Rupp, Markus},
  journal={IEEE Transactions on Signal Processing (TSP)}, 
  title={Likelihood Consensus and Its Application to Distributed Particle Filtering}, 
  year={2012},
  volume={60},
  number={8},
  pages={4334-4349},
  doi={10.1109/TSP.2012.2196697}}

@article{Tobin2017DomainRF,
  title={Domain randomization for transferring deep neural networks from simulation to the real world},
  author={Joshua Tobin and Rachel Fong and Alex Ray and Jonas Schneider and Wojciech Zaremba and P. Abbeel},
  journal={2017 IEEE/RSJ International Conference on Intelligent Robots and Systems (IROS)},
  year={2017},
  pages={23-30},
  url={https://api.semanticscholar.org/CorpusID:2413610}
}

@article{dellaert2017factor,
	title={Factor graphs for robot perception},
	author={Dellaert, Frank and Kaess, Michael and others},
	journal={Foundations and Trends in Robotics},
	volume={6},
	number={1-2},
	pages={1--139},
	year={2017},
}

@article{jacob2016coupling,
  title={Coupling of particle filters},
  author={Jacob, Pierre E and Lindsten, Fredrik and Sch{\"o}n, Thomas B},
  journal={arXiv},
  number={1606.01156},
  year={2016}
}

@article{srirama2024hrp,
  title={Hrp: Human affordances for robotic pre-training},
  author={Srirama, Mohan Kumar and Dasari, Sudeep and Bahl, Shikhar and Gupta, Abhinav},
  journal={arXiv},
  number={2407.18911},
  year={2024}
}

@article{damen2020epic,
  title={The epic-kitchens dataset: Collection, challenges and baselines},
  author={Damen, Dima and Doughty, Hazel and Farinella, Giovanni Maria and Fidler, Sanja and Furnari, Antonino and Kazakos, Evangelos and Moltisanti, Davide and Munro, Jonathan and Perrett, Toby and Price, Will and others},
  journal={IEEE Transactions on Pattern Analysis and Machine Intelligence},
  volume={43},
  number={11},
  pages={4125--4141},
  year={2020},
  publisher={IEEE}
}

@inproceedings{grauman2022ego4d,
  title={Ego4d: Around the world in 3,000 hours of egocentric video},
  author={Grauman, Kristen and Westbury, Andrew and Byrne, Eugene and Chavis, Zachary and Furnari, Antonino and Girdhar, Rohit and Hamburger, Jackson and Jiang, Hao and Liu, Miao and Liu, Xingyu and others},
  booktitle={Proceedings of the IEEE/CVF conference on computer vision and pattern recognition},
  pages={18995--19012},
  year={2022}
}

\end{document}